\pdfoutput=1

\documentclass[11pt]{article}

\usepackage[final]{acl}

\usepackage{times}
\usepackage{latexsym}

\usepackage[T1]{fontenc}

\usepackage[utf8]{inputenc}

\usepackage{microtype}

\usepackage{inconsolata}
\usepackage{amsmath}
\usepackage{hyperref}
\usepackage{url}
\usepackage{subfigure}
\usepackage{graphicx}
\usepackage{multicol}
\usepackage{multirow}
\usepackage{booktabs}
\usepackage{wrapfig}
\usepackage{algorithm}
\usepackage{algpseudocode}
\usepackage{colortbl}
\usepackage{soul}
\usepackage{threeparttable}
\definecolor{Gray}{gray}{0.9}
\sethlcolor{Gray}
\usepackage{threeparttable}

\title{Compensate Quantization Errors: Make Weights Hierarchical to Compensate Each Other}

\author{Yifei Gao$^1$, Jie Ou$^1$, Lei Wang$^{2}$\footnotemark[1], Yuting Xiao$^3$, Zhiyuan Xiang$^4$, Ruiting Dai$^1$,  Jun Cheng$^2$\\
    $^1$University of Electronic Science and Technology of China  \\ $^2$Shenzhen Institutes of Advanced Technology, Chinese Academy of Sciences \\ 
    $^3$Beijing Normal University\\
    $^4$Tianjin University\\
    \texttt{yilei.jin123@gmail.com, lei.wang1@siat.ac.cn} \\
         }

\begin{document}
\maketitle

\renewcommand{\thefootnote}{\fnsymbol{footnote}}
  \footnotetext[1]{Corresponding author.} 
  
\begin{abstract}
Emergent Large Language Models (LLMs) use their extraordinary performance and powerful deduction capacity to discern from traditional language models. However, the expenses of computational resources and storage for these LLMs are stunning, quantization then arises as a trending conversation. To address accuracy decay caused by quantization, two streams of works in post-training quantization methods stand out. One uses other weights to compensate existing quantization error, while the other transfers the quantization difficulty to other parts in the model. Combining both merits, we introduce Learnable Singular value Increment (LSI) as an advanced solution. LSI uses Singular Value Decomposition to extract singular values of the weights and make them learnable to help weights compensate each other conditioned on activation. Incorporating LSI with existing techniques, we achieve state-of-the-art performance in diverse quantization settings, no matter in weight-only, weight-activation or extremely low bit scenarios. By unleashing the potential of LSI, efficient finetuning on quantized model is no longer a prohibitive problem.
\end{abstract}

\section{Introduction}

Large language models (LLMs) have garnered significant attention for their remarkable performance across a wide range of downstream tasks and their ability to exhibit emergent behavior~\citep{gpt4,llama}. Furthermore, their prowess in understanding natural language and deductive reasoning can be extended to multimodal domains through alignment training~\citep{embodiedgpt,lvlm-ehub,meta-transformer}.
Nevertheless, the training and upkeep of such LLMs are highly resource-intensive, with many GPUs being able to support only a single model or parts of one. Quantization, as a central paradigm in this field, emerges as a solution that addresses both memory footprint and computational challenges.

Quantization methods are typically divided into two categories based on the quantization period. Quantization-Aware Training (QAT)~\citep{llmqat} involves tuning the model during training to optimize its compatibility with quantization. Although QAT can yield superior results compared to Post-Training Quantization (PTQ), the significant computational costs associated with the training process are a notable challenge. Consequently, PTQ methods have gained widespread acceptance and become increasingly popular in recent times.

Within the PTQ field, there are various methodologies to explore. For instance, the GPTQ series~\citep{obq,gptq,spqr,owq} employ unquantized weights to gradually offset the quantization errors introduced by previously quantized weights. In this paper, for the sake of convenience in reading, all references to "\textit{weight}" pertain to the \textit{weight matrices} in the linear layers of the model. They achieve this by solving a Lagrange equation to obtain a new Hessian matrix to update. On the other hand, methods like SmoothQuant~\citep{smoothquant} and OmniQuant~\citep{omniquant} focus on altering the distribution of weights and activations to mitigate the challenges of quantization. Meanwhile, through extensive research and experiments, it has been revealed that a significant portion of errors proposed during the quantization process are caused by a small number of outliers with distinctive weight values. Hence, several studies~\citep{spqr,outlier,outlier-plus,owq} concentrate on reducing or mitigating these outliers to minimize the disruption caused by their presence.

After thorough research, we have summarized the reasons for the success of previous methods. For the GPTQ series, in most cases, the weights obtained through manual calculation are not different from those obtained by the simple uniform quantization method, detailed in Sec~\ref{sec:quantization technique}. However, in some later stages of the quantization phase, GPTQ can alter the inherent hierarchy of the original weights, approximating the weights to other quantization intervals, to achieve a globally optimized solution. For methods like SmoothQuant, the concept of transferring the difficulty of weight quantization is relatively easy to understand. For example, if the equation $1.3 \times 15.4 \approx 20$ is going to be quantized through rounding, quantizing $1.3$ to $1$ would require changing $15.4$ to $20$ to maintain the original value. However, if the two multipliers in the equation are scaled to $4.3 \times 4.7 \approx 20$ and then quantized, it would only be necessary to quantize it to $4 \times 5$ to keep the original value unchanged. In this way, the difficulty of quantization is greatly simplified. Both types of methods have their own advantages, but they cannot both enjoy the benefits of the other. Additionally, approaches like the GPTQ series have drawbacks such as long quantization times.


In this paper, we demonstrate that a good quantization method performs: (1) Transformation of the quantization difficulty of weights and activations; (2) Hierarchical change of some weights to fit the global optimum; (3) Data-free in the PTQ setting; (4) A small amount of quantization time consumption; (5) Inference efficiency (mixed-precision is not allowed). Based on the requirements mentioned above, we introduce a novel technique called \textit{Learnable Singular value Increment} (LSI), which can effectively meet all the aforementioned requirements simultaneously. Unlike QAT, LSI exclusively focuses on training singular values of the weights, which constitute less than 0.1$\%$ of the total weights. Through the incorporation of a smoothing technique, LSI further simplifies the quantization problem. Moreover, existing methods primarily focus on aligning the performance of quantized models with unquantized models. However, with LSI, we have the capability to fine-tune quantized models without compromising the overall capabilities of these models themselves.

After conducting thorough experiments, we have achieved state-of-the-art results across a wide range of quantization settings, while marking a significant breakthrough in the field of quantized model fine-tuning. Our contributions encompass the following key points:
\begin{itemize}
\item Introduction of an innovative technique, LSI, which promotes hierarchical organization of model weights, facilitating their adaptation to quantization settings without compromising the inherent capacity of the model.

\item Integration of LSI with established smoothing techniques, effectively addressing the outlier issue and determining optimal transformation scales for quantization.

\item Demonstrating the applicability of LSI in fine-tuning quantized models under few-shot conditions, with the fine-tuning results showing significant improvements.
\end{itemize}

\section{Related Works}

\subsection{Quantization Methods}
\textbf{Weight-Only Quantization.} Previously proposed methods have primarily focused on weight-only quantization, where the emphasis is on converting the weight matrices of the model into low-bit representations. This approach allows for significant reductions in computational resources when storing and distributing models. For instance, a model with 30 billion parameters can be stored using a memory of as little as 20GB. However, in this setup, a significant portion of quantization errors arises from high-magnitude activations, often referred to as outliers. Many works~\citep{spqr,owq} have attempted to address this issue through mixed precision quantization while maintaining acceptable results. Nevertheless, this approach can introduce hardware inefficiencies and lead to increased inference time. Other methods, such as AWQ~\citep{awq}, employ more sophisticated scaling strategies, dividing weights into different channels, and exclusively incorporating quantization scales and zero points.

\textbf{Weight-Activation Quantization.} Weight-activation quantization involves the quantization of both model weights and activations. In the widely used self-attention setting, reducing the activation precision from 16 bits to a lower level results in a significant memory and time improvement, often of a squared magnitude.

In many cases, residual networks are interposed between layers, maintaining uniform and flat weights but leading to imbalanced and occasionally polarized activation. The quantization of activation is notably more challenging than that of weights, primarily because the outliers in activation are often substantially different from typical weight values.

\subsection{Quantization Techniques}
\label{sec:quantization technique}
\textbf{Uniform Quantization.} Although non-uniform quantization usually outperforms uniform quantization, it falls short in the dependency of some specialized devices~\citep{ant}. In contrast, uniform quantization is a more practical and feasible approach that can be efficiently executed on regular hardware devices. Uniform quantization discretizes a high-precision value into low-bit levels. Typically, we use the uniform quantization function $Q$ to transform a float linear weight matrix $W$ (commonly used in LLMs) to $k$ bits integer matrix $\tilde{W}$ as follows:
\begin{equation}
\footnotesize
\label{eq:uniform-quantization}
    \mathbf{\tilde{W}} = Q(\mathbf{W},k,s_{h},z) = Clamp(\lfloor \frac{\mathbf{W}}{s_{h}} \rceil + z, 0, 2^{k}-1),
\end{equation}
where $s_{h}$ and $z$ are corresponding shift and zero-point, respectively.

\textbf{Smooth}. The smooth technique is a method that involves transforming weights or activations by transferring certain magnitudes between them while maintaining their mathematical equivalence. Given a magnitude factor represented as a scaling matrix $diag(\mathbf{s_{c}})$, activations $\mathbf{X}$, and the final output $\mathbf{Y}$, the transformation can be driven as:
\begin{equation}
\footnotesize
\label{eq:smooth}
    \mathbf{Y} = \mathbf{X}\mathbf{W} + \mathbf{B} = [\underbrace{(\mathbf{X}-\delta)\oslash s_{c}}_{\tilde{\mathbf{X}}}]\cdot[\underbrace{s_{c}\odot \mathbf{W}}_{\tilde{\mathbf{W}}}] + [\underbrace{\mathbf{B} + \delta\mathbf{W}}_{\tilde{\mathbf{B}}}]
\end{equation}
where $\tilde{\mathbf{X}}, \tilde{\mathbf{W}}$ and $\tilde{\mathbf{B}}$ are equivalent activation, weight, and bias, respectively. $\oslash$ and $\odot$ are elementwise division and multiplication, respectively.


\section{Our Methods}

\subsection{Inspiration}
Our inspiration is drawn from the idea of compressing images into smaller sizes, where Singular Value Decomposition (\textbf{SVD}) is employed, and from the training method LoRA~\citep{lora}, which focuses on training sub-matrices rather than the entire weight matrix. In the context of quantization, we can perceive a quantized model as a training objective after quantization has been applied to the original model. The goal of the training process (quantization process) is to make the linear weights in the model discrete enough to meet the specified n-bit quantization setting. Building upon this concept, our goal is to enable the model parameters to autonomously adapt to the quantization process in a data-independent condition. In other words, we aim to have the weights automatically adjust to an appropriate magnitude that aligns with the specified n-bit setting without training on large datasets. We believe that by achieving this level of self-adjustment, we can attain better results in the context of quantization.

However, irrespective of how meticulously we design the quantization strategy, quantization errors are an inherent part of the process. It is crucial to recognize that not all differences resulting from quantization errors are detrimental. As mentioned in GPTQ~\citep{gptq}, when group-wise updates are made to the Hessian matrix during quantization, performance is nearly equivalent to updating the Hessian matrix one element at a time, as errors tend to compensate for each other throughout the procedure.

\begin{figure}
\includegraphics[width=0.5\textwidth]{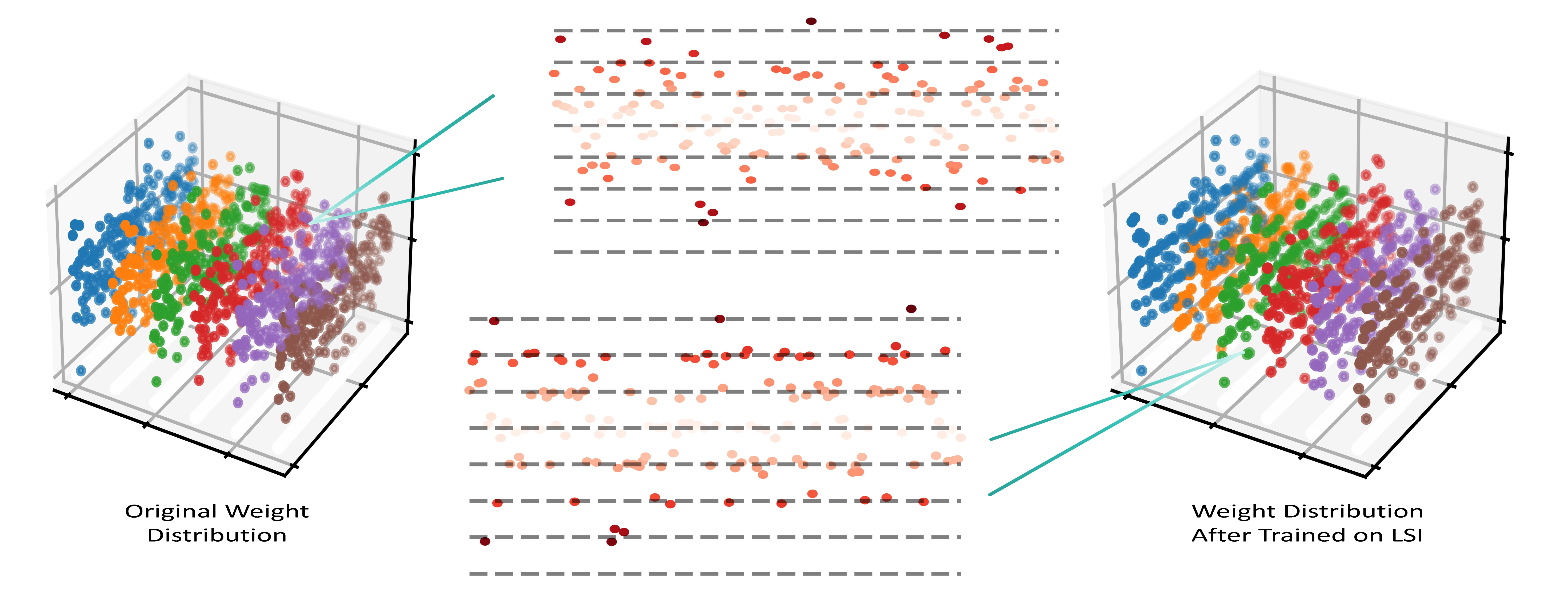}
\centering
\vspace{-1em}
    \caption{\textbf{Weight Distribution Comparison} between original weights and weights trained after LSI.}
    \label{fig:weight_distri}
    \vspace{-1.5em}
\end{figure}

Incorporating both of these fundamental insights, the Learnable Singular Value Increment (LSI) method emerges as a solution to address both training and performance issues in the context of quantization. It is noteworthy that our approach differs fundamentally from previous methods in the insight into quantization. While other methods primarily aim to reduce quantization errors, our approach makes use of these errors as a constructive element in the quantization process. We introduce weight disturbance by LSI to directly cause some "errors" to help the linear weights change their original magnitude to get the global optimum while not disturbing the original distribution that much. During quantization, LSI organizes weights hierarchically, grouping them into sets that closely resemble the specified discrete weight values, as presented in Fig. \ref{fig:weight_distri}.

\subsection{Learnable Singular Value Increment (LSI)}
\label{sec:LSI}

We now drive LSI optimization beginning with SVD. For a single linear layer, a weight matrix $\mathbf{W}^{a\times b}$ (we default $a \times b$) can be decomposed into three sub-matrices, $\mathbf{U}^{a \times a}, \mathbf{S}^{a}$ (only contain diagonal weights), $\mathbf{V_h}^{a \times b}$. Combining with uniform quantization, we assume that when we fix $\mathbf{U}$ and $\mathbf{V_h}$, there exists an optimal sub-matrix $\mathbf{S}^{\prime}$ (full matrix irrelevant to SVD) after QAT-like training. So we can get
\begin{equation}
\footnotesize
\label{eq:optimal-S}
    \mathbf{\tilde{W}} = \mathbf{U} \odot \mathbf{S}^{\prime} \odot \mathbf{V_h},
\end{equation}
which can largely satisfy the quantization setting and result in the least quantization error throughout the input. It means that given input $\mathbf{X}$, the output $\mathbf{Y}$, the minimum quantization error $\mathbf{E}_{min}$ conditioned on $\mathbf{S}^{\prime}$ can be expressed as:

\begin{equation}
\footnotesize
\label{eq:min-error}
    \mathbf{E}_{min} = \mathbf{Y} - \mathbf{X} \odot \mathbf{\tilde{W}} \\
    = \mathbf{Y} - \mathbf{X} \odot \mathbf{U} \odot \mathbf{S}^{\prime} \odot \mathbf{V_h}.
\end{equation}

In the equation provided above, once we identify the minimum error $\mathbf{E}_{min}$ and fix the quantization method $\mathbf{Q}$ along with variables $s$ and $z$, we can make $\mathbf{S}^{\prime}$ solvable. However, It is important to note that uniform quantization alone does not yield the optimal solution. Our goal is to introduce slight changes to the weight distribution, making the weights hierarchical to better conform to the uniform quantization setting. The change should be aware of the input since not all weights share the same importance during inference. 

Training the entire weight matrix is computationally intensive, and not all errors will have a negative impact, as mentioned previously. In our method, LSI introduces an additional variable called \textbf{Learnable Singular value Increment} $\mathbf{I}^{\prime}$ that is added on the original singular value to slightly change the weight distribution of original weight $\mathbf{W}$. In this setting, the quantized weight $\mathbf{\tilde{W}}$ can be obtained by:
\begin{equation}
\footnotesize
\begin{aligned}
\label{eq:LSI-quantization}
    & Q(\mathbf{W},k,s_{h},z,\mathbf{I}^{\prime}) \\
    & =  Clamp(\lfloor \frac{\mathbf{\mathbf{U} \odot 
    diag(\mathbf{S} + \mathbf{I}^{\prime}) \odot \mathbf{V_h}}} {s_{h}} \rceil + z, 0, 2^{k}-1), \\
    & \mathbf{\tilde{W}} = ( Q(\mathbf{W},k,s_{h},z,\mathbf{I}^{\prime}) -z ) s_{h}.
\end{aligned}
\end{equation}

And we optimize the corresponding equation to find the optimal $\mathbf{I}^{\prime}$ with linear function $\mathcal{F}$:
\begin{align}
\arg \min_{\mathbf{I}^{\prime}} \lvert\lvert \mathcal{F}(\mathbf{W},\mathbf{X}) - \mathcal{F}(\mathbf{\tilde{W}},\mathbf{X})) \rvert\rvert. 
\label{eq:objective}
\end{align}
Otherwise, if we shift our focuses on $\mathbf{S}^{\prime}$ (the optimal quantization sub-matrix), $\mathbf{I}^{\prime}$ is to make:
\begin{equation}
\footnotesize
\begin{aligned}
\label{eq:i_to_S_}
    \arg \min_{\mathbf{I}^{\prime}} \lvert\lvert \mathbf{U} \odot \mathbf{S}^{\prime} \odot \mathbf{V_h} -  Q(\mathbf{W},k,s_{h},z,\mathbf{I}^{\prime}) -z ) s_{h} \rvert\rvert.
\end{aligned}
\end{equation}
$\mathbf{I}^{\prime}$ is typically represented as a 1-D matrix, we cannot solely rely on LSI to perfectly align and significantly reduce errors. Instead, our approach deliberately introduces $\mathbf{I}^{\prime}$ as "errors" to facilitate minor disturbances in original weight distribution to compensate for each other from a global perspective. It's noteworthy that even if we train the quantized model layer by layer, our method can still largely achieve globally optimal results in the end. Due to the introduction of weight perturbation, later layers can largely compensate for the errors missed by earlier layers, which can be seen in Table~\ref{tab:lsi_finetuning}. This simplifies the training procedure while working towards the desired goal of error compensation.

In our experiments, however, we observed that for the group-wise setting, where different and delicate scaling scales are applied to various groups, LSI alone faces challenges in learning a set of optimal parameters that can balance all groups effectively. We will further discuss this issue in Sec~\ref{sec:ablation study}. To address this issue, we introduced an additional small square matrix, around the dimensions of $100 \times 100$ to $600 \times 600$, adding at the beginning of the diagonal matrix, specifically on the first $n$ rows and the first $n$ columns. This addition is made once the $diag(\mathbf{S} + \mathbf{I}^{\prime})$ component has been computed in Equation~\ref{eq:LSI-quantization}. We believe that the most prominent values are influenced by the most significant singular values, while the relatively smaller singular values have less impact on high-magnitude values. Therefore, the introduction of this additional square matrix helps achieve a better balance in the group-wise setting.

\subsection{Smooth and Clipping}
\label{sec:LET}
Techniques like \textit{Smooth} and \textit{Clipping} are powerful methods that find extensive use in various applications. \textit{Smooth} is effective at transferring quantization challenges to make the quantization of weights more manageable, while \textit{Clipping} is instrumental in addressing outlier issues. By incorporating these techniques with recent advancements, such as \textit{Learnable Weight Clipping (LWC)} and \textit{Learnable Equivalent Transformation (LET)} proposed in OmniQuant~\citep{omniquant}, LSI can significantly reduce quantization errors and achieve remarkable performance gains.

In the linear layer, LET is to make $diag(s_{c})$ in Eq. ~\ref{eq:smooth} learnable. Additionally, in the attention operation, LET introduces learnable parameter $diag(s_{a})$ to smooth the query $\mathbf{Q}$ and key $\mathbf{K}$, which can be represented as:
\begin{equation}\label{eq:LET-attn}
\footnotesize
\begin{aligned}
    \mathbf{P} 
    &=\mathrm{Softmax} (\mathbf{Q} \mathbf{K} ^{T})\\
    &=\mathrm{Softmax} ((\underbrace{\mathbf{Q}\oslash s_a}_{\tilde{\mathbf{Q}}}) (\underbrace{s_a\odot\mathbf{K} ^{T}}_{\tilde{\mathbf{K}}^T})).
\end{aligned}    
\end{equation}
On the other hand, LWC is to make upper and lower boundaries in quantization function $Q$ learnable in Eq.~\ref{eq:uniform-quantization} instead of fixed $2^{k}$.

\begin{table*}[t]
\footnotesize
    \setlength{\tabcolsep}{5pt}
    \centering
    \caption{\textbf{WikiText2 perplexity of Weight-only quantization results in OPT models}.}
    \begin{tabular}{llccccccc}
        \hline
          \multicolumn{2}{l}{\textbf{OPT / PPL}$\downarrow$} & 125M  & 1.3B & 2.7B & 6.7B & 13B & 30B & 66B\\  \hline
        FP16 & - & 27.65  & 14.63 & 12.47 & 10.86 & 10.12 & 9.56 
        & 9.34\\ 
        \hline
        \multirow{4}{*}{\shortstack{W2A16-g128}} 
         & GPTQ \citep{gptq} &  597.66  & 115.16 & 61.59 & 20.18 & 21.36 &  12.71 &   82.10 \\
         & AWQ \citep{awq} & 251.84  & 47.97 & 28.50 & 16.20 & 14.32 &  12.31 &  14.54 \\
         & OmniQuant \citep{omniquant} & 75.43  & 23.95 & 18.13 & 14.43 & 12.94 & 11.39 & 30.84 \\
         & \cellcolor{Gray}\textbf{Ours}  & \cellcolor{Gray}\textbf{56.17}  & \cellcolor{Gray}\textbf{22.59} & \cellcolor{Gray}\textbf{17.65} & \cellcolor{Gray}\textbf{14.23} & \cellcolor{Gray}\textbf{12.75} & \cellcolor{Gray}\textbf{11.30} & \cellcolor{Gray}\textbf{29.66}  \\
        \hline
        \multirow{4}{*}{\shortstack{W2A16-g64}} 
         & GPTQ \citep{gptq} &  204.40 & 49.58 & 29.37 & 16.81 & 16.65 &  11.87 &   356.01\\
         & AWQ \citep{awq} & 124.18  & 29.78 & 20.64 & 14.63 & 13.28 &  11.59 &   12.74\\
         & OmniQuant \citep{omniquant} & 62.56  & 21.40 & 16.76 & 13.57& 12.33 & 11.00 & 10.59\\
         & \cellcolor{Gray}\textbf{Ours}  & \cellcolor{Gray}\textbf{50.94}  & \cellcolor{Gray}\textbf{21.09} & \cellcolor{Gray}\textbf{16.69} & \cellcolor{Gray}\textbf{13.51} & \cellcolor{Gray}\textbf{12.25} & \cellcolor{Gray}\textbf{10.95} & \cellcolor{Gray}\textbf{10.56} \\ \hline
        \multirow{4}{*}{\shortstack{W3A16}} 
         & GPTQ \citep{gptq} &  53.05  & 21.17 & 16.83 & 15.09 & 11.73 &  10.30 &  14.42\\
         & AWQ \citep{awq} & 69.43 & 28.01 & 263.10 & 15.13 & 20.09 &  35.74 &  4.5e3 \\
         & OmniQuant \citep{omniquant} & 35.66  & 16.68 & 13.80 & 11.65 & 10.87 & 10.00 & 9.83\\
         & \cellcolor{Gray}\textbf{Ours}  & \cellcolor{Gray}\textbf{32.19} & \cellcolor{Gray}\textbf{16.24} & \cellcolor{Gray}\textbf{13.44} & \cellcolor{Gray}\textbf{11.46} & \cellcolor{Gray}\textbf{10.66} & \cellcolor{Gray}\textbf{9.96}  & \cellcolor{Gray}\textbf{9.79}\\ \hline
         \multirow{4}{*}{\shortstack{W4A16}} 
         & GPTQ \citep{gptq} & 31.43 & 15.56 & 12.82 & 11.41 & 10.31 & 9.63 & 9.55\\
         & AWQ \citep{awq} & 32.28 & 15.49 & 12.93 & 11.30 & 10.39 &  9.77 &   9.61 \\
         & OmniQuant \citep{omniquant} & 29.45 & 15.04 & 12.76 & 11.03 & 10.30 & 9.65 & 9.65 \\
         & \cellcolor{Gray}\textbf{Ours} & \cellcolor{Gray}\textbf{28.86} & \cellcolor{Gray}\textbf{15.00} & \cellcolor{Gray}\textbf{12.71} & \cellcolor{Gray}\textbf{11.00} &  \cellcolor{Gray}\textbf{10.24} & 
         \cellcolor{Gray}\textbf{9.63}  & 
         \cellcolor{Gray}\textbf{9.50}\\ 
         \hline
    \end{tabular}
    \label{tab:opt weight only wiki}
\end{table*}

\begin{table*}[tb]
\footnotesize
\centering
\caption{ \textbf{Weight-activation quantization results of OPT Models.} We report perplexity on three datasets: WikiText2 (WIKI), Pen Treebank (PT), and C4. RPTQ indicates the data from RPTQ~(\cite{rptq}) paper, which keeps the output of LN and SoftMax as 8-bit. RPTQ$^*$ represents reproducing RPTQ with our setting that quantizes all activation into low-bit except keeping the softmax output at full precision. OPT-66B results can be found in supplementary material.}
\label{tab:opt_weight_activation_ppl}
{
\begin{tabular}{llcccccccccccc}
\hline
\multicolumn{2}{l}{\textbf{OPT / PPL}$\downarrow$}  & \multicolumn{3}{c}{OPT-6.7b} & \multicolumn{3}{c}{OPT-13b} & \multicolumn{3}{c}{OPT-30b}\\ \hline
\multicolumn{2}{l}{Task}      & WIKI     & PT      & C4      & WIKI    & PT      & C4      & WIKI    & PT      & C4    \\ \hline
FP16   & -  & 10.86    & 13.09   & 11.74   & 10.13   & 12.34   & 11.20   & 9.56    & 11.84   & 10.69   \\
\hline
\multirow{5}{*}{\shortstack{W6A6}} & SmoothQuant \citep{smoothquant} & 11.34  & 13.82  & 12.14   & 10.56  & 12.76   & 11.40   & 9.67   & 12.01   & 10.81   \\
& RPTQ \citep{rptq} & 11.19  & 13.98  & 12.08   & 11.00  & 15.23   & 11.68   & 10.22   & 14.95   & 11.73 \\
& RPTQ$^*$ & 10.96  & 13.24  & 11.86   & 10.25  & 12.60   & 11.31   & 9.60   & 12.23   & 10.83 \\
& OmniQuant \citep{omniquant} & 10.96  & 13.20  & 11.81   & 10.21  & 12.47   & 11.27   & 9.62   & \textbf{11.92}   & 10.76 \\
& \cellcolor{Gray}\textbf{Ours} & \cellcolor{Gray}\textbf{10.91}  & \cellcolor{Gray}\textbf{13.19}  & \cellcolor{Gray}\textbf{11.80}   & \cellcolor{Gray}\textbf{10.19}  & \cellcolor{Gray}\textbf{12.45}   & \cellcolor{Gray}\textbf{11.27}   & \cellcolor{Gray}\textbf{9.60}   & \cellcolor{Gray}{11.93}   & \cellcolor{Gray}\textbf{10.75} \\   
\hline
\multirow{5}{*}{\shortstack{W4A4}} & SmoothQuant \citep{smoothquant} & 1.8e4  & 1.4e4  & 1.5e4   & 7.4e3  & 6.5e3   & 5.6e3   & 1.2e4   & 7.8e3   & 8.3e3  \\
 & RPTQ  \citep{rptq}  & 12.00    & 15.17   & 12.85   & 12.74   & 15.76   & 14.71   & 11.15   & 14.11   & 13.48    \\
 & RPTQ$^*$    & 17.83    & 25.10   & 19.91   & 16.45  & 23.01   & 16.80   & 11.50   & 14.87   & 12.81  \\
  & OmniQuant  \citep{omniquant}  & 12.24    & 15.54   & 13.56    & 11.65  & 15.89   & 13.46   & 10.60   & 13.75   & 11.89  \\
& \cellcolor{Gray}\textbf{Ours}   & \cellcolor{Gray}\textbf{11.82}  & \cellcolor{Gray}\textbf{14.86}  & \cellcolor{Gray}\textbf{13.10}   & \cellcolor{Gray}\textbf{11.10}  & \cellcolor{Gray}\textbf{15.16}   & \cellcolor{Gray}\textbf{12.81}   & \cellcolor{Gray}\textbf{10.29}   & \cellcolor{Gray}\textbf{13.32}   & \cellcolor{Gray}\textbf{11.64}   \\
\hline
\end{tabular}
}
\end{table*}

\begin{table*}[t]
    \setlength{\tabcolsep}{3pt}
    \footnotesize
    \centering
    \caption{\textbf{Weight-activation quantization results of LLaMA Models.} We report our perplexity results on WikiText2 and C4, along with accuracy (for multi-choices tasks, we report our accuracy norm) of 6 zero-shot tasks compared with other baselines.(In W6A6 settings, layers in LLaMA can not train on LSI somehow)}
    \begin{threeparttable}
    \begin{tabular}{lllcccccccc}
        \hline
          \multicolumn{1}{l}{\textbf{LLaMA / Acc}$\uparrow$} & \#Bits & Method  & Wiki & C4 & PIQA  & ARC-e & Arc-c & HS$^1$ & WG$^2$ & \textbf{Avg.} \\  \hline
        \multirow{7}{*}{LLaMA-7B} &
        FP16 & -  & -  & - & 77.47  & 52.48 & 41.46 & 73.00 & 67.07 & 62.30  \\
        & W4A4 & SmoothQuant \citep{smoothquant} & - & - & 49.80 & 30.40  & 25.80 & 27.40 & 48.00 & 36.28 \\
        & W4A4 & LLM-QAT & - & - & 51.50 & 27.90 & 23.90 & 31.10 & 51.90 & 37.26 \\
        & W4A4 & LLM-QAT+SQ & - & - & 55.90 & 35.50 & 26.40 & 47.80 & 50.60 & 43.26 \\
        & W4A4 & OmniQuant \citep{omniquant} & 11.26 & 14.51 & 66.15 & 45.20 & 31.14 & 56.44 & 53.43 & 50.47 \\
        & \cellcolor{Gray}\textbf{W4A4} & \cellcolor{Gray}\textbf{Ours} 
        & \cellcolor{Gray}\textbf{11.02}  & \cellcolor{Gray}\textbf{13.77}
        & \cellcolor{Gray}\textbf{67.90} & \cellcolor{Gray}\textbf{47.43} & \cellcolor{Gray}\textbf{31.91} & \cellcolor{Gray}\textbf{57.51} & \cellcolor{Gray}\textbf{56.27} & \cellcolor{Gray}\textbf{52.20} \\
        \hline
        \multirow{5}{*}{LLaMA-13B} &
        FP16 & - & -  & - & 79.10  & 59.89 & 44.45 & 76.21 & 70.31 & 65.99  \\
        & W4A4 & SmoothQuant \citep{smoothquant} & - & - & 61.04 & 39.18 & 30.80 & 52.29 & 51.06 & 46.87 \\
        & W4A4 & OmniQuant \citep{omniquant} & 10.87 & 13.78 & 69.69 & 47.39 & 33.10 & 58.96 & 55.80 & 53.05 \\
        &\cellcolor{Gray} W4A4 & \cellcolor{Gray}\textbf{Ours}
        & \cellcolor{Gray}\textbf{10.68} & \cellcolor{Gray}\textbf{12.84} 
        & \cellcolor{Gray}\textbf{69.69} & \cellcolor{Gray}\textbf{47.43} & \cellcolor{Gray}\textbf{33.61} & \cellcolor{Gray}\textbf{62.37} & \cellcolor{Gray}\textbf{59.59} & \cellcolor{Gray}\textbf{54.54} \\
        \hline
        \multirow{5}{*}{LLaMA-30B} &
        FP16 & - & - & - & 80.08  & 58.92 & 45.47 & 79.21 & 72.53 & 67.24 \\
        & W4A4 & SmoothQuant \citep{smoothquant} & - & - & 58.65 & 35.53 & 27.73 & 35.56 & 48.06 & 41.11 \\
        & W4A4 & OmniQuant \citep{omniquant} & 10.33 & 12.49 & 71.21 & 49.45 & 34.47 & 64.65 & 59.19 & 55.79 \\
        & \cellcolor{Gray}W4A4 & \cellcolor{Gray}\textbf{Ours} 
        & \cellcolor{Gray}\textbf{10.20} & \cellcolor{Gray}\textbf{12.12} 
        & \cellcolor{Gray}\textbf{72.90} & \cellcolor{Gray}\textbf{49.45} & \cellcolor{Gray}\textbf{36.43} & \cellcolor{Gray}\textbf{65.98} & \cellcolor{Gray}\textbf{60.22} & \cellcolor{Gray}\textbf{57.00} \\
        \hline
    \end{tabular}
    \begin{tablenotes}    
         \scriptsize 
         \item[1] `HS'  stands for HellaSwag.
         \item[2]  `WG' stands for WinoGrande.
      \end{tablenotes} 
    \end{threeparttable}
    \label{tab:llama_weight_activation}
\end{table*}

\section{Experiments}

\subsection{Settings}

\textbf{Quantization.}
In line with the methodology outlined in~\citep{omniquant}, our experiments cover both weight-only and weight-activation quantization settings. For the weight-only component, we employ channel-wise weight quantization at INT4/INT3/INT2 bit levels. In the weight-activation setting, we utilize quantization settings of w6a6/w4a4, where `w' and `a' signify \textit{weight}  and \textit{activation}, respectively. In cases where groups are divided, with each group having a distinct set of quantization parameters, we use `g' to represent the group name. Furthermore, we adhere to the original setup, keeping the Softmax part in float32, as this helps mitigate excessive disturbance caused by self-attention layers during the inference process. We also inherit the enhanced acceleration nature of OmniQuant in INT3/INT2 settings on CUDA. Please see~\citep{omniquant} for more details.

\textbf{Training.} Given that LSI effectively satisfies both smoothing and shifting techniques, we initialize the scaling and shifting parameters using well-trained parameters from~\citep{omniquant}. We then train a set of LSI parameters based on this initial setup. Singular values can introduce significant variations in the distribution of weights, so we maintain a low learning rate at 2e-4. We employ the AdamW~\citep{adamw} optimizer with a weight decay of 0 to optimize our parameters. All the data used in our training was collected from WikiText2~\citep{wikitext2}. Notably, the training process is quite fast, with larger models requiring fewer epochs. For instance, in the `w4a16g128' setting, the \textit{OPT-30B} model only needs to be trained for 2 epochs on a dataset with 32 samples. All techniques proposed before were included during the whole quantization procedures. Additionally, without groups, we all set square matrix dimension $n=200$. But with group-wise scaling, we test several dimensions to validate the effectiveness of the increment square matrix, which we will discuss in Sec~\ref{sec:ablation study}. For the finetuning, we select PTB~\citep{ptb}, where the perplexity of our baselines on it is significantly higher than others. We only train the last two layers of models with epochs around 10 to 40 on 128 PTB samples, which is very fast to implement.

\textbf{Models.} We conduct evaluation on two popular baselines for generalization, LLaMA(7-30B)~\citep{llama}, OPT(125M-66B)~\citep{opt}. For more details about our test results, please see the supplementary materials.

\textbf{Evaluation.} Our evaluation for perplexity is mainly focused on WikiText2~\citep{wikitext2}, PTB~\citep{ptb}), C4~\citep{c4}. Furthermore, following previous works, we also evaluate several zero-shot tasks in weight-activation quantization setting, including PIQA~\citep{piqa}, ARC~\citep{arc}, and HellaSwag~\citep{arc}. Samples of datasets we evaluated obey the  GPTQ~\citep{gptq} settings. For accuracy tasks, lm-eval-harness~\citep{evaluation} is employed for all zero-shot tasks.

\textbf{Baselines.} For weight-only quantization, we choose previously state-of-the-art works, GPTQ~\citep{gptq}, AWQ~\citep{awq} and recently advanced work OmniQuant~\citep{omniquant} as baselines. For weight-activation quantization, both QAT and PTQ methods are included, containing SmoothQuant~\citep{smoothquant}, RPTQ~\citep{rptq}, QAT~\citep{llmqat} and Omniquant~\citep{omniquant}. Following SmoothQuant \citep{smoothquant}, we do not change the per-channel quantization strategy for weights and the per-tensor quantization strategy for activation.

\subsection{Weight-only Quantization Results}
In this section, we mainly demonstrate the results of the OPT series  without group-wise scaling in w3a16 and w4a16 settings on WikiText2, as shown in Table~\ref{tab:opt weight only wiki}, while w3a16 and w4a16 group-wise results are shown in Sec~\ref{sec:ablation study}. Our full results can be found in the supplementary material. As exhibited in these tables, prominent progress can be seen in various settings. LSI provides effective and powerful solutions for the quantization of smaller LLMs and w2 settings while helping further improve the performance in more sophisticated settings.

\subsection{Weight-Activation Quantization Results}
In the context of weight-activation quantization, we have successfully improved several metrics over the original OmniQuant \citep{omniquant}, as presented in Table~\ref{tab:opt_weight_activation_ppl}, and achieved enhanced performance across various tasks with the LLaMA families, as shown in Table~\ref{tab:llama_weight_activation}. Specifically, in W6A6 settings, we generally observe slightly better results, and in W4A4 settings, we significantly outperform existing methods.

\begin{table*}[tb]
\centering
\caption{ \textbf{Weight-only} quantization results of OPT Models (125m-2.7b) in W3A16g128 and W4A16g128 settings with different $k$.}
\label{tab:opt_weight_activation_125m-2.7b}
\footnotesize
{
\begin{tabular}{llcccccccccccc}
\hline
\multicolumn{2}{l}{\textbf{OPT / PPL}$\downarrow$}  & \multicolumn{3}{c}{OPT-125m} & \multicolumn{3}{c}{OPT-1.3b} & \multicolumn{3}{c}{OPT-2.7b}\\ \hline
\multicolumn{2}{l}{Task}      & WIKI     & PT      & C4      & WIKI    & PT      & C4      & WIKI    & PT      & C4    \\ \hline
FP16   & -  & 27.65    & 32.54   & 24.60   & 14.63   & 16.96   & 14.72  & 12.47    & 15.11   & 13.16   \\
\hline
\multirow{5}{*}{\shortstack{W3A\\16g128}} & GPTQ \citep{gptq} &   39.24  & 45.17  &  30.08   &  16.47  & 19.90   &  16.47   &  13.69   &  17.06   & 14.54 \\
& AWQ \citep{awq} & 36.74  & 44.07  & 30.39   & 16.32  & 19.59   & 16.27   & 13.58   & 16.52   & 14.19 \\
& OmniQuant \citep{omniquant} & 32.25  & 40.76  & 29.34   & 15.72  & 19.06   & 16.11   & 13.18   & 16.29   & \textbf{14.15} \\
& \cellcolor{Gray}\textbf{Ours k100} & \cellcolor{Gray}\underline{31.63}  & \cellcolor{Gray}\underline{40.74}  & \cellcolor{Gray}\underline{29.21}
& \cellcolor{Gray}\underline{15.68}  & \cellcolor{Gray}\underline{18.99}   & \cellcolor{Gray}\underline{16.11}
& \cellcolor{Gray}\underline{13.17}   & \cellcolor{Gray}\textbf{16.27}   & \cellcolor{Gray}{14.17} \\
& \cellcolor{Gray}\textbf{Ours k200} & \cellcolor{Gray}\textbf{31.06}  & \cellcolor{Gray}\textbf{39.84}  & \cellcolor{Gray}\textbf{28.78}   
& \cellcolor{Gray}\textbf{15.64}  & \cellcolor{Gray}\textbf{18.95}   & \cellcolor{Gray}\textbf{16.09}   
& \cellcolor{Gray}\textbf{13.15}   & \cellcolor{Gray}\textbf{16.27}   & \cellcolor{Gray}\underline{14.16} \\   
\hline
\multirow{5}{*}{\shortstack{W4A\\16g128}} & GPTQ \citep{gptq} & 29.81  & 35.48  & 25.96   & 14.89 &  \underline{17.41} &  15.05   &  \textbf{12.52}   & 15.42 & 13.40  \\
 & AWQ  \citep{awq}  &  29.15    & 34.95  & 25.90   &  14.94  & 17.46 &  15.04   & 12.74   & 15.33   &  \underline{13.39}  \\
  & OmniQuant \citep{omniquant}   & 28.86    & 34.28   & 25.63    & 14.88  & \textbf{17.40}  & \underline{15.03}   & 12.65   & 15.28   & 13.38  \\
& \cellcolor{Gray}\textbf{Ours k100}   & \cellcolor{Gray}\underline{28.57}  & \cellcolor{Gray}\textbf{33.68}  & \cellcolor{Gray}\underline{25.51}   
& \cellcolor{Gray}\underline{14.87}  & \cellcolor{Gray}{17.42}   & \cellcolor{Gray}\textbf{15.02}   & \cellcolor{Gray}{12.64}   & \cellcolor{Gray}\textbf{15.26}   & \cellcolor{Gray}\textbf{13.38}   \\
& \cellcolor{Gray}\textbf{Ours k200}   & \cellcolor{Gray}\textbf{28.40}  & \cellcolor{Gray}\underline{34.21}  & \cellcolor{Gray}\textbf{25.45}   
& \cellcolor{Gray}\textbf{14.85}  & \cellcolor{Gray}{17.44}   & \cellcolor{Gray}\textbf{15.02}   
& \cellcolor{Gray}\underline{12.62}   & \cellcolor{Gray}\underline{15.28}   & \cellcolor{Gray}\textbf{13.38}   \\
\hline
\end{tabular}
}
\end{table*}

\begin{table*}[tb]
\centering
\caption{ \textbf{Weight-only} quantization results of OPT Models (6.7m-30b) in W3A16g128 and W4A16g128 settings. Here, we adopt $k=320$ in OPT-6.7b, $k=450$ in OPT-13b and $k=600$ in OPT-30b}
\label{tab:opt_weight_activation_6.7b-30b}
\footnotesize
{
\begin{tabular}{llcccccccccccc}
\hline
\multicolumn{2}{l}{\textbf{OPT / PPL}$\downarrow$}  & \multicolumn{3}{c}{OPT-6.7m} & \multicolumn{3}{c}{OPT-13b} & \multicolumn{3}{c}{OPT-30b}\\ \hline
\multicolumn{2}{l}{Task}      & WIKI     & PT      & C4      & WIKI    & PT      & C4      & WIKI    & PT      & C4    \\ \hline
FP16   & -  & 10.86    &  13.08   & 11.74   & 10.12   &  12.33   &  11.19  &  9.56    &  11.84   & 10.69   \\
\hline
\multirow{5}{*}{\shortstack{W3A\\16g128}} & GPTQ \citep{gptq} &   11.65  & 14.24  &  12.48   &   10.35  & 12.84   &  11.58   &   9.73   &  12.54   & 10.91 \\
& AWQ \citep{awq} & 11.41  & 13.98  &  12.30  &  10.68  &  12.87   & 11.61   & 9.85  &  66.68   & 10.96 \\
& OmniQuant \citep{omniquant} & 11.27  & 13.77  & 12.31   & 10.47  & 12.96   & 11.63   & 9.79  & 12.19   & 10.98 \\
& \cellcolor{Gray}\textbf{Ours k200} & \cellcolor{Gray}{11.26}  & \cellcolor{Gray}{13.77}  & \cellcolor{Gray}{12.31}
& \cellcolor{Gray}{10.45}  & \cellcolor{Gray}{12.94}   & \cellcolor{Gray}{11.63}
& \cellcolor{Gray}{9.79}   & \cellcolor{Gray}{12.17}   & \cellcolor{Gray}{10.98} \\
& \cellcolor{Gray}\textbf{Ours k320-600} & \cellcolor{Gray}{11.26}  & \cellcolor{Gray}{13.76}  & \cellcolor{Gray}{12.31}   
& \cellcolor{Gray}{10.45}  & \cellcolor{Gray}{12.95}   & \cellcolor{Gray}{11.62}   
& \cellcolor{Gray}{9.76}   & \cellcolor{Gray}{12.19}   & \cellcolor{Gray}{10.98} \\   
\hline
\multirow{5}{*}{\shortstack{W4A\\16g128}} & GPTQ \citep{gptq} & 10.93  & 13.21  & 11.87  & 11.26 & 12.42 & 12.46   &  9.58   & 11.89 &  10.74  \\
 & AWQ  \citep{awq}  &  10.93    & 13.28  &  11.87   & 10.21  & 12.46 & 11.28   &  9.59   & 11.90   &  10.75  \\
  & OmniQuant  \citep{omniquant}  & 10.96    & 13.25   & 11.85    & 10.20  & 12.46 & 11.29  & 9.62   & 11.94   & 10.75  \\
& \cellcolor{Gray}\textbf{Ours k200}   & \cellcolor{Gray}{10.95}  & \cellcolor{Gray}{13.25}  & \cellcolor{Gray}{11.85}   
& \cellcolor{Gray}{10.19}  & \cellcolor{Gray}{12.47}   & \cellcolor{Gray}{11.29}   
& \cellcolor{Gray}{9.61}   & \cellcolor{Gray}{11.95}   & \cellcolor{Gray}{10.74}   \\
& \cellcolor{Gray}\textbf{Ours k320-600}   & \cellcolor{Gray}{10.94}  & \cellcolor{Gray}{13.24}  & \cellcolor{Gray}{11.85}   
& \cellcolor{Gray}{10.19}  & \cellcolor{Gray}{12.46}   & \cellcolor{Gray}{11.29}   
& \cellcolor{Gray}{9.61}   & \cellcolor{Gray}{11.93}   & \cellcolor{Gray}{10.75}   \\
\hline
\end{tabular}
}
\end{table*}

\begin{table*}[tb]
\centering
\caption{ \textbf{LSI-only} quantization results}
\label{tab:lsi_only_results}
\footnotesize
{
\begin{tabular}{llcccccccccccc}
\hline
\multicolumn{2}{l}{\textbf{PPL}}& \multicolumn{2}{c}{LLaMA-7b} & \multicolumn{3}{c}{OPT-2.7b} & \multicolumn{3}{c}{OPT-6.7b}\\ \hline
\multicolumn{2}{l}{Task}      & WIKI      & C4      & WIKI    & PT      & C4      & WIKI    & PT      & C4    \\ \hline

\multirow{2}{*}{\shortstack{W3A16g128}} & GPTQ \citep{gptq} &   6.55   &  7.85   &   13.69  & 17.06   &  14.54   &   11.65   &   14.24   & 12.48 \\
& \cellcolor{Gray}\textbf{Ours} 
& \cellcolor{Gray}{6.25}  & \cellcolor{Gray}{7.91}
& \cellcolor{Gray}{13.70} & \cellcolor{Gray}{17.35}   & \cellcolor{Gray}{14.82}
& \cellcolor{Gray}{11.75}   & \cellcolor{Gray}{14.87}   & \cellcolor{Gray}{13.05} \\  
\hline

\multirow{2}{*}{\shortstack{W4A16}} & GPTQ \citep{gptq} & 6.13  & 7.43  & 12.82 &  15.94 & 13.75   &  11.41   & 13.75 &   12.15  \\
& \cellcolor{Gray}\textbf{Ours}   
& \cellcolor{Gray}{5.95} & \cellcolor{Gray}{7.47}  
& \cellcolor{Gray}{12.76}  & \cellcolor{Gray}{16.15}   & \cellcolor{Gray}{14.06}   
& \cellcolor{Gray}{11.27}   & \cellcolor{Gray}{13.93}   & \cellcolor{Gray}{12.33}   \\
\hline
\end{tabular}
}
\end{table*}

\begin{table}[tb]
\centering
\scriptsize
\caption{ \textbf{Finetuning} by LSI}
\label{tab:lsi_finetuning}
{
\begin{tabular}{llcccccccccccc}
\hline
\multicolumn{2}{l}{\textbf{PPL}}& \multicolumn{3}{c}{LLaMA-7b}\\ \hline
\multicolumn{2}{l}{Task}  & PT    & WIKI      & C4  \\ \hline
\multirow{4}{*}{\shortstack{W3A\\16g128}} 
& RTN &   37.37   &  7.01   &  8.62 \\
& \cellcolor{Gray}\textbf{RTN w/ LSI} 
& \cellcolor{Gray}{35.58}  & \cellcolor{Gray}{6.91} & \cellcolor{Gray}{8.52}  \\  
& OmniQuant \citep{omniquant} & 33.45   &  {6.15}   & {7.75}\\
& \cellcolor{Gray}\textbf{Omni w/ LSI} 
& \cellcolor{Gray}{30.69}  & \cellcolor{Gray}{6.16} & \cellcolor{Gray}{7.77} \\  
\hline
\multicolumn{2}{l}{\textbf{PPL}}& \multicolumn{3}{c}{LLaMA-30b}\\ \hline
\multirow{4}{*}{\shortstack{W4A16}} 
& RTN & 17.15 & 4.57 & 6.34\\
& \cellcolor{Gray}\textbf{RTN w/ LSI}   
& \cellcolor{Gray}{17.06} & \cellcolor{Gray}{4.55}   & \cellcolor{Gray}{6.32}\\
& OmniQuant \citep{omniquant} & 16.48  & {4.25}  & {6.11} \\
& \cellcolor{Gray}\textbf{Omni w/ LSI}   
& \cellcolor{Gray}{16.46} & \cellcolor{Gray}{4.26}  & \cellcolor{Gray}{6.12}  \\
\hline
\end{tabular}
}
\end{table}

\subsection{Ablation Study}\label{sec:ablation study}
In our extensive ablation studies, we investigated the effectiveness of LSI and the impact of the additional increment square matrix. Table~\ref{tab:opt_weight_activation_125m-2.7b} indicates that LSI is remarkably beneficial when the model size is relatively small, and adjusting the value of $k$ indeed provides certain advantages. However, as the volume of the models increases, particularly with the influence of group-wise scaling, the benefits brought by LSI and its corresponding matrix diminish, and can even facilitate the overfitting problem as observed in Table~\ref{tab:opt_weight_activation_6.7b-30b}. Meanwhile, the introduction of the adding matrix can result in some affinity to some specific datasets, which may be a kind of overfitting problem.

We posit that this diminishing impact might be attributed to group-wise scaling, which discretizes the entire weight matrix into different parts, resulting in incoherence within the weight matrix. This implies that the compensation for quantization error is restricted to individual groups, disregarding the integrated nature of the entire matrix. Considering the coherence of the weight matrix, singular values influence the distribution of the entire weight matrix rather than a singular part.

On the other hand, LSI alone can achieve competitive performance, however, it suffers severely from bias. Without the transformation of quantization difficulty, LSI obtains relatively good performance through significant overfitting on a specific dataset, as seen in Table~\ref{tab:lsi_only_results}.

\subsection{Finetuning of LSI}
As illustrated in Sec~\ref{sec:ablation study}, LSI alone has a grave problem of overfitting. After investigations, we find that LSI can help bridge the gaps caused by previous layers, which means that even only employing LSI in the last several layers, there are still some overfitting problems. However, if we use this property to quickly finetune a model on a specific dataset, this weakness turns into an advantage. In finetuning part, we first employ LSI to change the original weight distribution and then transfer the quantization difficulty using smooth techniques.  Through our experiments, in nearly all settings, employing LSI only on the last several layers of a LLM can also result in improved performance on a specific dataset without largely compromising other abilities, as shown in Table~\ref{tab:lsi_finetuning}. To test the generalization of LSI finetuning, we prepare \textit{OmniQuant} and \textit{Round To Nearest (RTN)} as baseline quantization strategies and only employ LSI in the last two layers. LSI can also satisfy other baselines, as we have tried to use GPTQ baselines and replace the last several layers quantized with our techniques, it also works well.

\subsection{Other Issues}
\textbf{Inference Speed.} Overall, our method does not introduce additional inference time. Because after training, we integrate LSI into the original weights to alter them, and then quantize them to the specified bit. For the introduced smooth technology, only LWC will cause a minor delay in inference. However, for LET, the transfer of weight scaling in LET is integrated into the norm function of each layer. After training, it directly scales the gain in the Layernorm function of the original model, so there will be no impact during inference. For detailed inference speeds, one can refer to~\citep{omniquant}. Our inference speed is essentially identical to theirs.

\textbf{Best Results.} Our best results were not obtained by using OmniQuant's parameters as the initialization. During the training process, we found that random initialization followed by a longer training period could potentially yield better results. However, due to the instability of random initialization and the loss incurred by the extended quantization time, we did not use random initialization in our experiments.

\section{Limitations}
\subsection{Overfitting Problem} \label{sec:overfitting_problem}
Despite the introduction of additional parameters constituting an extremely small portion compared to the overall parameter volume, their significance is substantial. Aligned with our philosophy, the redistribution of model weights should be cognizant of activations, which leads to the overfitting problem. As demonstrated in our supplementary material, there is a trade-off between achieving improvements in perplexity on one dataset and a cost associated with the other. 

In our experiments, we observed that LSI can indeed significantly reduce the loss caused by quantization. However, the elimination of loss does not always guarantee a corresponding performance improvement. This phenomenon is further substantiated through the training procedure, as discussed in our supplementary material.

Generally speaking, LSI tends to achieve superior results by largely aligning with one dataset. In the early stages, when the overall error is substantial, conforming to one dataset can substantially enhance coherence and restore the original capacity of the model. However, as errors gradually diminish, a boundary is encountered. Given that errors are unavoidable in quantization settings, beyond this boundary, LSI exhibits overfitting problems.

\subsection{Hard to Train}
As discussed in Sec~\ref{sec:overfitting_problem}, there is a boundary in training. But for different models with different volumes, these boundaries do not display a stable paradigm, so it is needed to train with different epochs to gradually get closer to the optimal performance. But in general, with model volume growing, less training is needed. For example, in the W4A16 setting, 5 epochs are needed to train the \textit{OPT-13B} on a 128-sample dataset, while only 1 epoch is needed for \textit{OPT-30B} on a 32-sample dataset, which takes less than 1.5 hours. For more details, please see our supplementary material. 

On the other hand, when implementing only LSI, the entire process becomes somewhat precarious. As our goal is to align the quantized weight distribution, LSI can assist weights in stepping over their original magnitude span. In this process, significant fluctuations are quite common, and even with the incorporation of smoothing techniques, this phenomenon cannot be entirely avoided.

\section{Conclusion}
We introduce LSI to adjust model weights to conform to quantization settings. Through integration with established techniques, our approach attains state-of-the-art performance across diverse quantization settings. LSI imparts a hierarchical structure to model weights, enhancing adaptability to quantization parameters without compromising training efficiency. Leveraging attributes of LSI allows effective finetuning of a quantized model on various datasets. Notably, during inference, LSI introduces no additional parameters and preserves the hardware efficiency inherited from OmniQuant.

\section*{Acknowledgements}
This work was supported in part by the Guangdong Major Project of  Basic and Applied Basic Research (2023B0303000016), Shenzhen Technology Project (JCYJ20220531095810023), National Natural Science Foundation of China (U21A20487, 61976143), CAS Key Technology Talent Program, Colleges and Universities Key Laboratory of Intelligent Integrated Automation (No.201502),  CAS Key Laboratory of Human-Machine Intelligence-Synergy Systems, Shenzhen Institutes of Advanced Technology, Chinese Academy of Sciences (2014DP173025), Guangdong-Hong Kong-Macao Joint Laboratory of Human-Machine Intelligence-Synergy Systems (2019B121205007).

\bibliography{acl}

\clearpage
\newpage

\appendix

\section{Appendix}
\label{sec:appendix}

In this appendix, we provide further details as follows:
\begin{itemize}
    \item Sec. \ref{sec:train-analysis}: Training procedure analysis of the local sensitivity information, and when optimal results can be achieved using \textbf{Learnable Single value Increment} (LSI).
    \item Sec. \ref{sec:full-results}: Showcases the complete results for OPT, LLaMA-1 models.
\end{itemize}

\subsection{Training Procedure Analysis}\label{sec:train-analysis}
\begin{figure}[ht]
\includegraphics[width=0.4\textwidth]{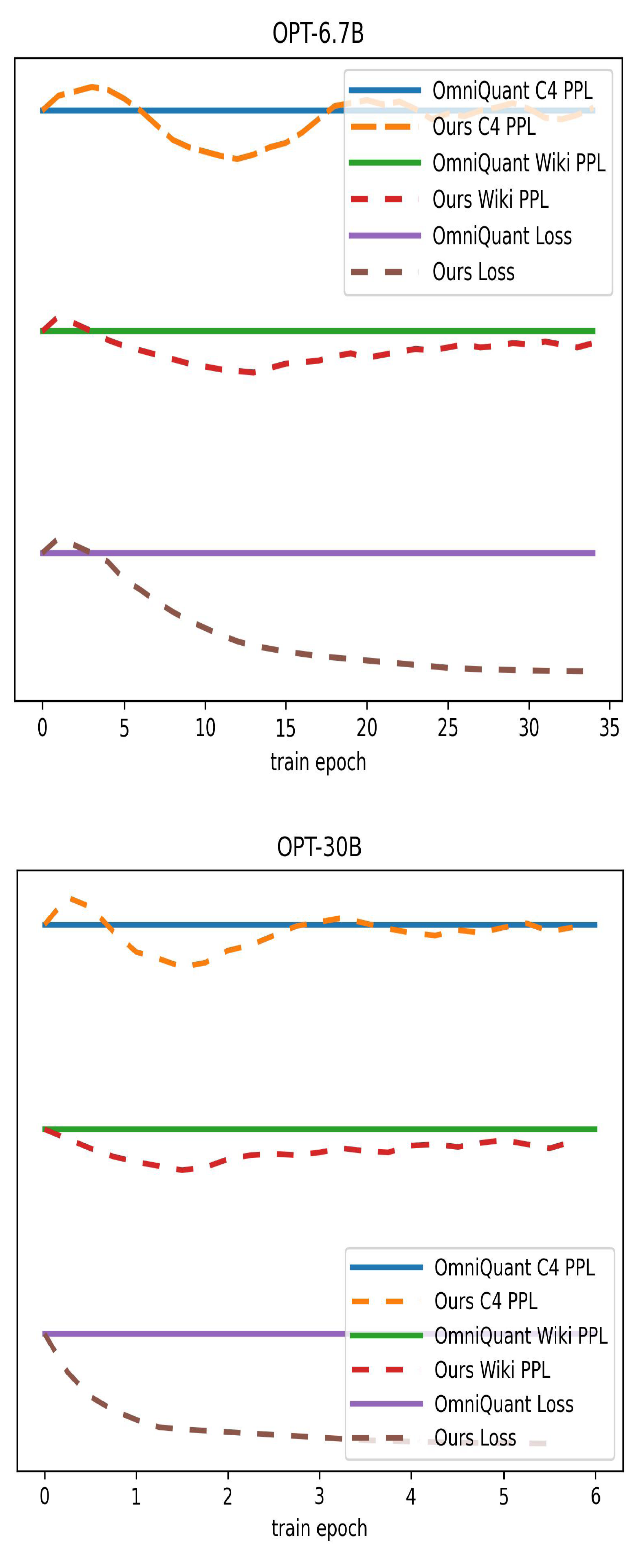}
\centering
\vspace{-1em}
    \caption{\textbf{Training details about LSI} on OPT-6.7B and OPT-30B in the setting of W4A16g128. The magnitude is not drawn in scale. Since OPT-30B is really sensitive to train epochs, in the OPT-30B part, we make one train epoch containing 24 samples, and we train it from 8 samples to 144 samples at the interval of 8 samples.}
    \label{fig:train_proces}
    \vspace{-1.5em}
\end{figure}

In this section, our focus centers on elucidating the training procedure for our techniques. Fig~\ref{fig:train_proces} illustrates that exhaustive training does not consistently yield the optimal outcome. Notably, when trained on OPT-30B, a mere 32 samples suffice; however, even slight deviations within this range can induce noteworthy disturbances in performance. This poses difficulties in getting the optimal parameter.

LSI proves highly effective in mitigating quantization loss. Generally, LSI can achieve a reduction ranging from 20$\%$ to 40$\%$, surpassing the performance of OmniQuant. Nevertheless, it's crucial to note, as emphasized in our paper, that the reduction in loss does not necessarily translate to improved overall performance.

\begin{table}[b]
\scriptsize
\centering
\caption{ \textbf{Weight-activation quantization results} of OPT-66B. We test the results that only use LSI in the last several layers, and \textit{L} refers to the layer number implemented LSI. RPTQ$^*$ represents reproducing RPTQ with our setting that quantizes all activation into low-bit except keeping the softmax output at full precision.}
\label{tab:opt_66b_weight_activation_ppl}
{
\begin{tabular}{llccc}
\hline
\multicolumn{2}{l}{\textbf{OPT / PPL}$\downarrow$}  &  \multicolumn{3}{c}{OPT-66b}\\ \hline
\multicolumn{2}{l}{Task}      & WIKI     & PT      & C4  \\ \hline
FP16   & -  & 9.34    & 11.36  & 10.28 \\
\hline
\multirow{6}{*}{\shortstack{W4A4}} & SmoothQuant \citep{smoothquant} & 2.2e5  & 1.e5  & 1.8e5 \\
 & RPTQ  \citep{rptq} & 12.23   & 18.87   & 15.93   \\
 & RPTQ$^*$    & 11.16   & 13.73   & 11.78 \\
  & OmniQuant  \citep{omniquant} & 10.29    & 13.19   & 11.35 \\
& \cellcolor{Gray}\textbf{Ours L4}
& \cellcolor{Gray}\textbf{10.26}  & \cellcolor{Gray}\textbf{13.30}  & \cellcolor{Gray}\textbf{11.31}  \\
& \cellcolor{Gray}\textbf{Ours}  
& \cellcolor{Gray}\textbf{10.21}  & \cellcolor{Gray}\textbf{13.08}  & \cellcolor{Gray}\textbf{11.26}  \\
\hline
\end{tabular}
}
\end{table}

\subsection{Full Results}\label{sec:full-results}
In this section, we provide a comprehensive presentation of our results across various datasets to complement the main paper. Specifically, the results include:
\begin{itemize}
    \item OPT-66B results on W4A4 setting (Table \ref{tab:opt_66b_weight_activation_ppl}).
    \item Wiki perplexity with weight-only quantization in the LLaMA families (Table \ref{tab:llama_weight_only_results}).
    \item PTB perplexity with weight-only quantization in OPT families (Table \ref{tab:opt_weight_only_ptb}).
    \item C4 perplexity with weight-only quantization in OPT families (Table \ref{tab:opt_weight_only_c4}).
\end{itemize}

\begin{table*}[ht]
\centering
\caption{ \textbf{Weight-only} quantization results of LLaMA-1 Models (7B-30B) in settings without group-wise scaling. The LLaMA families exhibit insensitivity to scaling and weight distribution; only handling outliers can make a noticeable difference. In OmniQuant, they solely employ LWC instead of LWC+LET, meaning all progress is achieved through outlier elimination. In this context, we observed that LSI does not bring about significant improvements in group-wise settings. For LLaMA-7B and LLaMA-13b, there is a slight increment with changes around 0.01 degree of perplexity on average. However, for LLaMA-30B, the increment is nearly negligible. Therefore, the implementation of LSI is not considered necessary in group-wise scaling on LLaMA families. But we also release our checkpoint of LLaMA-7b and LLaMA-13b on those settings, so anyone can check for examination.}

\label{tab:llama_weight_only_results}
\footnotesize
{
\begin{tabular}{llccccccccc}
\hline
\multicolumn{2}{l}{\textbf{LLaMA / PPL}$\downarrow$}  & \multicolumn{2}{c}{LLaMA-7B} & \multicolumn{2}{c}{LLaMA-13B} & \multicolumn{2}{c}{LLaMA-30B}\\ \hline
\multicolumn{2}{l}{Task}      & WIKI     & C4      & WIKI     & C4      & WIKI    & C4    \\ \hline
FP16   & -  & 10.86    & 11.74   & 10.12  &  11.19  &  9.56   & 10.69   \\
\hline
\multirow{3}{*}{\shortstack{W2A16}} 
& GPTQ \citep{gptq} & 2.1e3  & 689.13 & 5.5e3 & 2.5e3 & 499.75 &  169.80   \\
& OmniQuant \citep{omniquant} & 15.47  & 24.89  & 13.21  & 18.31 & 8.71  & 13.89 \\
& \cellcolor{Gray}\textbf{Ours} & \cellcolor{Gray}\textbf{12.91}  & \cellcolor{Gray}\textbf{17.90}  & \cellcolor{Gray}\textbf{9.08}
& \cellcolor{Gray}\textbf{12.36}  & \cellcolor{Gray}\textbf{8.45}   & \cellcolor{Gray}\textbf{11.96}\\  
\hline
\multirow{4}{*}{\shortstack{W3A16}} 
& GPTQ \citep{gptq} &  8.06  & 9.49  & 6.76 & 8.16  & 5.84  &  7.29   \\
& AWQ \citep{awq} & 11.88  & 13.26 & 7.45  & 9.13  & 10.07   &  12.67 \\
& OmniQuant \citep{omniquant} & 6.49  & 8.19  & 5.68  & 7.32 & 4.74  & 6.57  \\
& \cellcolor{Gray}\textbf{Ours} & \cellcolor{Gray}\textbf{6.38}  & \cellcolor{Gray}\textbf{8.17}  & \cellcolor{Gray}\textbf{5.65}
& \cellcolor{Gray}7.33  & \cellcolor{Gray}\textbf{4.69}   & \cellcolor{Gray}6.58\\  
\hline
\multirow{4}{*}{\shortstack{W4A16}} 
& GPTQ \citep{gptq} & 6.13  & 7.43 & 5.40 & 6.84  & 4.48 &  6.20   \\
& AWQ \citep{awq} & 6.08  & 7.52 & 5.34  & 6.86  & 4.39   & 6.17 \\
& OmniQuant \citep{omniquant} & 5.86  & 7.34 & 5.21  & 6.76 & 4.25  & 6.11 \\
& \cellcolor{Gray}\textbf{Ours} & \cellcolor{Gray}\textbf{5.84}  & \cellcolor{Gray}\textbf{7.32}  & \cellcolor{Gray}\textbf{5.20}
& \cellcolor{Gray}\textbf{6.75}  & \cellcolor{Gray}\textbf{4.24}   & \cellcolor{Gray}\textbf{6.11}\\  
\hline
\end{tabular}
}
\end{table*}

\begin{table*}[!ht]
    \setlength{\tabcolsep}{5pt}
    \small
    \centering
    \caption{\cellcolor{Gray}\textbf{PTB perplexity of Weight-only quantization results in OPT models}.}
    \begin{tabular}{llccccccc}
        \hline
          \multicolumn{2}{l}{\textbf{OPT / PPL}$\downarrow$} & 125M  & 1.3B & 2.7B & 6.7B & 13B & 30B & 66B \\  \hline
        FP16 & - & 32.54 & 16.96  & 15.11 & 13.08  & 12.33 & 11.84 & 11.36\\ 
        \hline
        \multirow{4}{*}{\shortstack{W2A16\\g128}} 
         & GPTQ \citep{gptq} & 655.17  & 130.88 & 61.36 & 25.24 & 20.46 & 15.15 & 323.23\\
         & AWQ \citep{awq} & 263.88  & 71.87 & 43.15 & 19.49 & 17.61 & 14.92 & 19.33 \\
         & OmniQuant \citep{omniquant} & 126.49  & 34.33 & 25.28 & 18.92 & 16.74 & 14.51 & 139.17\\
         & \cellcolor{Gray}\textbf{Ours}  & \cellcolor{Gray}\textbf{92.60} & \cellcolor{Gray}\textbf{32.26} & \cellcolor{Gray}\textbf{24.39} & \cellcolor{Gray}\textbf{18.71} & \cellcolor{Gray}\textbf{16.44} & \cellcolor{Gray}\textbf{14.27} & \cellcolor{Gray}\textbf{116.21}\\
        \hline
        \multirow{4}{*}{\shortstack{W2A16\\g64}} 
         & GPTQ \citep{gptq} & 245.28  & 55.61 & 36.12 & 19.45 & 17.02 & 14.05 & 88.92\\
         & AWQ  \citep{awq} & 143.18 & 41.19 & 25.08 & 18.00 & 15.83 & 14.92 & 15.72\\
         & OmniQuant \citep{omniquant} & 112.10  & 30.36 & 22.63 & 17.58 & 15.70 & 13.98 & 13.51\\
         & \cellcolor{Gray}\textbf{Ours}  & \cellcolor{Gray}\textbf{81.40} & \cellcolor{Gray}\textbf{29.17} & \cellcolor{Gray}\textbf{22.51} & \cellcolor{Gray}\textbf{17.55} & \cellcolor{Gray}\textbf{15.55} & \cellcolor{Gray}\textbf{13.90} & \cellcolor{Gray}\textbf{13.47} \\
         \hline
        \multirow{4}{*}{\shortstack{W3A16}} 
         & GPTQ \citep{gptq} & 34.05  & 27.39 & 15.94 & 13.75 & 13.71 & 12.54 & 21.16\\
         & AWQ  \citep{awq} & 80.73  & 33.20 & 224.11 & 18.46 & 35.45 & 66.68 & 3.4e3\\
         & OmniQuant \citep{omniquant} & 45.29  & 20.42 & 17.08 & 14.23 & 13.49 & 12.54 & 11.71\\
         & \cellcolor{Gray}\textbf{Ours}  & \cellcolor{Gray}\textbf{40.56} & \cellcolor{Gray}\textbf{19.85} & \cellcolor{Gray}\textbf{16.65} & \cellcolor{Gray}\textbf{14.02} & \cellcolor{Gray}\textbf{13.42} & \cellcolor{Gray}\textbf{12.48} & \cellcolor{Gray}\textbf{11.69} \\
         \hline
         \multirow{4}{*}{\shortstack{W4A16}} 
         & GPTQ \citep{gptq} & 37.75  & 18.23 & 15.94 & 13.75 & 12.58 & 11.98 & 11.58\\
         & AWQ \citep{awq} & 38.74  & 18.35 & 15.70 & 13.59 & 12.72 & 12.06 &  11.58\\
         & OmniQuant \citep{omniquant} & 34.94 & 17.80 & 15.52 & 13.41 & 12.62 & 11.95  & 11.86\\
         & \cellcolor{Gray}\textbf{Ours}  & \cellcolor{Gray}\textbf{34.83} & \cellcolor{Gray}\textbf{17.74} & \cellcolor{Gray}\textbf{15.43} & \cellcolor{Gray}\textbf{13.37} & \cellcolor{Gray}\textbf{12.55} & \cellcolor{Gray}\textbf{11.95}  & \cellcolor{Gray}\textbf{11.73} \\
         \hline
    \end{tabular}
    \label{tab:opt_weight_only_ptb}
\end{table*}

\begin{table*}[!ht]
    \setlength{\tabcolsep}{5pt}
    \small
    \centering
    \caption{\textbf{C4 perplexity of Weight-only quantization results in OPT models}. }
    \begin{tabular}{llcccccccc}
        \hline
          \multicolumn{2}{l}{\textbf{OPT / PPL}$\downarrow$} & 125M  & 1.3B & 2.7B & 6.7B & 13B & 30B & 66B\\  \hline
        FP16 & - & 24.60  & 14.72 & 13.16 & 11.74 & 11.19 & 10.69 & 10.69\\ 
        \hline
        \multirow{4}{*}{\shortstack{W2A16\\g128}} 
         & GPTQ \citep{gptq} & 597.66 & 60.88 & 33.83 & 18.55 & 16.34 & 12.89 & 598.81\\
         & AWQ \citep{awq} & 168.35  & 38.38 & 26.41 & 16.48 & 14.73 & 12.98 & 15.42\\
         & OmniQuant  \citep{omniquant} & 80.10  & 27.33 & 21.11 & 16.67 & 14.92 & 13.12 & 73.83\\
         & \cellcolor{Gray}\textbf{Ours}  & \cellcolor{Gray}\textbf{64.17} & \cellcolor{Gray}\textbf{25.76} & \cellcolor{Gray}\textbf{20.61} & \cellcolor{Gray}\textbf{16.28} & \cellcolor{Gray}\textbf{14.66} & \cellcolor{Gray}\textbf{13.00} & \cellcolor{Gray}\textbf{66.25}\\
        \hline
        \multirow{4}{*}{\shortstack{W2A16\\g64}} 
         & GPTQ \citep{gptq} & 133.51 & 31.31 & 23.23 & 16.24 & 14.48 & 12.24 & 58.60\\
         & AWQ \citep{awq} & 90.19  & 27.34 & 20.01 & 15.20 & 13.90 & 12.43 & 13.31\\
         & OmniQuant \citep{omniquant} & 64.01  & 23.71 & 19.16 & 15.44 & 14.16 & 12.80 & 12.13\\
         & \cellcolor{Gray}\textbf{Ours}  & \cellcolor{Gray}\textbf{56.22} & \cellcolor{Gray}\textbf{23.53} & \cellcolor{Gray}\textbf{19.03} & 
         \cellcolor{Gray}15.31 & 
         \cellcolor{Gray}13.97 
         & \cellcolor{Gray}\textbf{12.75}  
         & \cellcolor{Gray}\textbf{12.10}\\
         \hline
        \multirow{4}{*}{\shortstack{W3A16}} 
         & GPTQ \citep{gptq} & 37.75  & 19.45 & 13.75 & 15.67 & 12.28 & 11.34 & 13.68\\
         & AWQ \citep{awq} & 55.73  & 24.56 & 154.49 & 15.84 & 23.71 & 55.01 &  3.8e3\\
         & OmniQuant \citep{omniquant} & 32.17  & 17.10 & 14.93 & 12.78 & 12.13 &  11.37 &  10.82\\
         & \cellcolor{Gray}\textbf{Ours}  & \cellcolor{Gray}\textbf{30.20} & \cellcolor{Gray}\textbf{16.71} & \cellcolor{Gray}\textbf{14.59} & \cellcolor{Gray}\textbf{12.56} & 
         \cellcolor{Gray}12.14 & 
         \cellcolor{Gray}11.38  & 
         \cellcolor{Gray}\textbf{10.79}  \\
         \hline
         \multirow{4}{*}{\shortstack{W4A16}} 
         & GPTQ \citep{gptq} & 27.12  & 15.57 & 13.75 & 12.15 & 11.36 & 10.80 &  10.50\\
         & AWQ \citep{awq} & 27.64  & 15.65 & 13.71 & 12.04 & 11.42 & 10.83 &  10.41 \\
         & OmniQuant \citep{omniquant} & 26.36  & 15.28 & 13.58 & 11.97 & 11.41 & 10.80 & 10.63\\
         & \cellcolor{Gray}\textbf{Ours}  & \cellcolor{Gray}\textbf{26.02} & \cellcolor{Gray}\textbf{15.26} & \cellcolor{Gray}\textbf{13.52} & \cellcolor{Gray}\textbf{11.94} & 
         \cellcolor{Gray}11.37 
         & \cellcolor{Gray}\textbf{10.79}  
         & \cellcolor{Gray}\textbf{10.47} \\
         \hline
    \end{tabular}
    \label{tab:opt_weight_only_c4}
\end{table*}

\end{document}